\documentclass[10pt,letterpaper]{article}

\usepackage{cogsci}

\cogscifinalcopy 
 \usepackage[table,xcdraw]{xcolor}
\usepackage{graphicx}
\usepackage{listings}
\usepackage{booktabs}
\usepackage{multirow, tabularx}
\usepackage{amsmath,amssymb}
\usepackage{algorithm2e} 
\usepackage{pslatex}
\usepackage{apacite}
\usepackage{soul}
\usepackage{subcaption}
\usepackage{comment} 
\usepackage{csvsimple}
\usepackage{float} 

\usepackage[round]{natbib}
\usepackage{hyperref} 
\usepackage{comment}

\usepackage{colortbl} 
\usepackage{xcolor}   
\usepackage{multirow} 
\usepackage{graphicx} 
\usepackage{pgffor}   
\usepackage{tikz}     

\RestyleAlgo{ruled}

\title{Learning Efficient Recursive Numeral Systems via Reinforcement Learning}
\author{\large \bf
    Andrea Silvi \quad Jonathan Thomas  \quad Emil Carlsson \quad Devdatt Dubhashi \quad Moa Johansson \\ 
    Chalmers University of Technology and University of Gothenburg
}

\begin{document}

\maketitle

\renewcommand\thefootnote{} 
\footnotetext{Correspondence to: \texttt{silvi@chalmers.se}.}
\renewcommand\thefootnote{\arabic{footnote}} 

\begin{abstract}

It has previously been shown that by using reinforcement learning (RL), agents can derive simple approximate and exact-restricted numeral systems that are similar to human ones \cite{carlsson2021numeral}. However, it is a major challenge to show how more complex recursive numeral systems, similar to for example English, could arise via a simple learning mechanism such as RL.
Here, we introduce an approach towards deriving a mechanistic explanation of the emergence of efficient recursive number systems. We consider pairs of agents learning how to communicate about numerical quantities through a meta-grammar that can be gradually modified throughout the interactions. 
Utilising a slightly modified version of the meta-grammar of \cite{Hurford-1975}, we demonstrate that our RL agents, shaped by the pressures for efficient communication, can effectively modify their lexicon towards Pareto-optimal configurations which are comparable to those observed within human numeral systems in terms of their efficiency.

\textbf{Keywords:} efficient communication; reinforcement learning; numeral systems 

\end{abstract}

\section{Introduction}

While there is evidence to suggest that animals, young infants and adult humans possess a biologically determined, domain-specific representation of numbers and elementary arithmetic operations, only humans have a capacity for generating an infinite set of natural numbers, while all other species seem to lack such a capacity
\citep{hauser2002faculty, chomsky1982note, chomsky1986knowledge, dehaene1993development, dehaene1997numbers}. 
This unique capacity is central to many aspects of human cognition, including, of course, the development of
sophisticated mathematics. The fundamental mechanism underlying this is the use of a finite symbolic system to represent arbitrarily large discrete numerical magnitudes i.e. the positive integers. 
In the context of AI however, there is little work on developing and changing representations of mathematical concepts, such as number systems, while most primarily concerns revising representations of logical theories \cite{bundy23}. 

In cognitive science, a recent influential body of work suggests language is shaped by a pressure for efficient communication which involves an information-theoretic tradeoff between cognitive load and informativeness \citep{Kemp2012, Gibson2016, Zaslavsky2019a}. This means that language is under pressure to be simultaneously informative, in order to support effective communication, while also being simple, to minimize cognitive load. Exact and approximate numeral systems were studied in \cite{Xu2020}, while a mechanistic explanation of how such schemes could arise was proposed in \cite{carlsson2021numeral} via reinforcement learning in signaling games. However, these studies were limited to simpler exact and approximate numeral systems and do not cover more complex systems capable of representing arbitrary numerical quantities.

How does one explain the origins and development of numerical systems capable of generating expressions for arbitrary numerical quantities? \cite{chomsky2008phases} hypothesises that a fundamental operation called \emph{Merge} can give rise to the \emph{successor} function (i.e., every numerosity N has a unique successor, N + 1) in a set-theoretic fashion (1 = one, 2 = \{one\}, 3=\{one, \{one\}\}, ...) and that the capacity for discretely infinite natural numbers may
be derived from this. However, representations of numbers in natural languages do not reveal any straightforward trace of the successor function.

Another proposal is that \emph{Merge} is able to \emph{integrate} the two more primitive number systems mentioned above, an approximate number system for large numerical quantities and a system of precise representation of distinct small numbers. A natural way to achieve this is via a \emph{grammar} that has two primitive components corresponding to these notions: a \emph{base} for large approximate numerical values and \emph{digits} for precise representations of small quantities. An example of a meta-grammar which covers this family of grammars is that proposed in the seminal work of \cite{Hurford-1975}. 



Such a grammar in turn would be subject to pressures to achieve a system of \emph{efficient communication} in the information-theoretic sense of \citep{Kemp2012, Gibson2016, Zaslavsky2019a}. Taking this line of thought, we investigate here how pressure for efficient representations would lead to the evolution of recursive numeral systems that are efficient in an information-theoretic sense in the context of artificial agents. We build on previous work by \citep{carlsson2021numeral} for approximate and exact restricted systems and extend their framework to consider recursive numeral systems representations, obtained via the use of \citet{Hurford-1975}'s meta-grammar. Tradeoffs in recursive numeral systems have been recently explored in \citet{denicRecursiveNumerals}, and we adapt their framework. However, a mechanistic system to learn such efficient numeral systems was not provided in \citet{denicRecursiveNumerals}. We provide such a system in a setting where two agents interact to communicate numerosity concepts. Our agents' emergent languages tend closer to the Pareto frontier in terms of the tradeoff between average morphosyntactic complexity and lexicon size, providing a mechanistic explanation to the work in \citet{denicRecursiveNumerals} and complementary to the one of \citet{carlsson2021numeral} for the other classes of numeral systems.


\begin{figure}[htbp]
    \centering
    \includegraphics[width=0.45\textwidth]{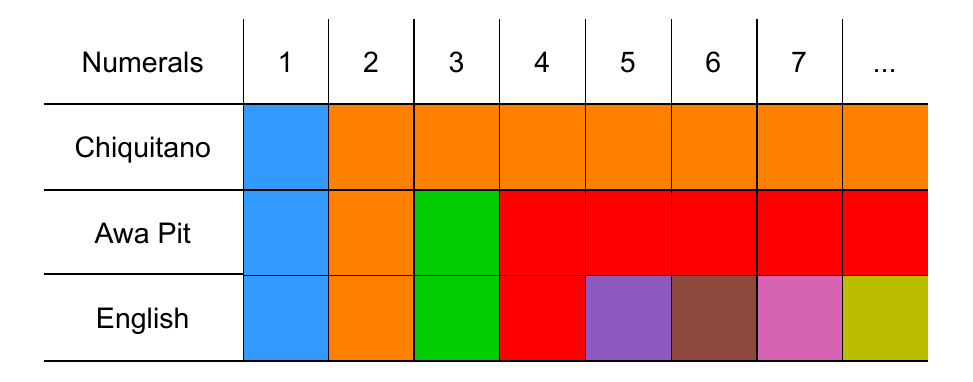}
    \caption{Three distinct numeral systems from three differing languages are shown. These are the approximate numerals used in \textit{Chiquitano}, the exact-restricted numerals used in \textit{Awa Pit} and the recursive numerals used in \textit{English}. Colors indicate how words are assigned to numeral concepts.}
    \label{fig:num_systems_differences}
\end{figure}

\section{Efficiency of Recursive Numeral Systems}

\begin{figure}[t]
    \centering
    \includegraphics[scale=0.55]{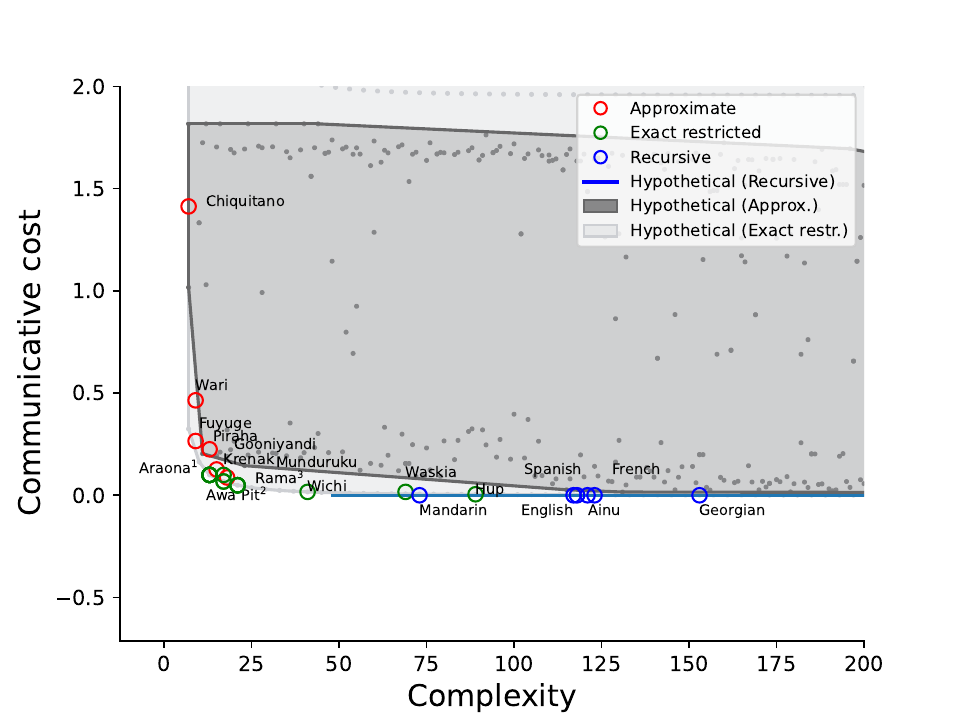}
    \caption{Reproduction of Figure 4b from \cite{Xu2020}, showing that while restricted numeral systems seem to optimize the simplicity/informativeness tradeoff (here are plotted the complexity and communicative cost, their opposite), recursive numeral systems (plotted as blue dots) do not, as they lie far away from the Pareto-optimal recursive numeral system (the left-most point of the blue line).}
    \label{fig:inform_simplicity_tradeoff}
\end{figure}

There is a large literature on the need for efficient communication as a  unifying principle that shapes human languages across several domains, see e.g. \citet{Kemp2012, regier2015word, Gibson2016, xu2016historical, kemp2018semantic, zaslavsky2018efficient, Xu2020, steinert2021quantifiers, mollica2021forms, zaslavsky2021let, uegaki2023informativeness, yin24acl} and the review \cite{Kemp2018} which covers several domains. A domain where this has been particularly well studied is that of numeral systems. Under this information-theoretic framework, \citet{Xu2020} argued that numeral systems, including approximate, exact restricted and recursive systems, support efficient communication (examples of these systems are shown in Figure \ref{fig:num_systems_differences}). While this argument is compelling for exact and approximate systems, it was pointed out by \citet{denicRecursiveNumerals} that for recursive systems, the picture is more complicated. The complexity measure of \citet{Xu2020} depends both on the number of lexicalized terms and number of rules in the grammar. As noted by \citet{denicRecursiveNumerals}, while exact and restricted systems lie very close to the information-theoretic frontier in the tradeoff between simplicity and informativeness, recursive systems seem to lie off the frontier, as shown in Figure \ref{fig:inform_simplicity_tradeoff}. The definition of complexity used by \citet{Xu2020} might not be well suited to recursive systems because these systems always maximize informativeness, since they can express any numeral. Hence, for recursive systems there is not a clear tradeoff between informativeness and this measure of complexity and hence recursive systems do not lie on the frontier. Instead, \citet{denicRecursiveNumerals} argue that there is (at least) one more pressure shaping systems of recursive numerals in addition to simplicity and informativeness, namely \emph{average morphosyntactic complexity}. Thus, since recursive systems achieve perfect informativeness, the appropriate tradeoff is along two axes: the number of lexicalized terms and the average morphosyntactic complexity. \citet{denicRecursiveNumerals} show that recursive systems found in human languages optimize a tradeoff between these two measures. A natural question that emerges from this line of argument is: what mechanistic processes can optimize for an efficient tradeoff along these dimensions?

 Like \citet{denicRecursiveNumerals} we first show how the Pareto frontier of the number of lexicalised terms and average morphosyntactic complexity can be estimated using a genetic algorithm optimising a grammar for number systems like the one of \cite{Hurford-1975, Hurford1987LanguageAN}. We then introduce a slightly modified Hurford-grammar better suited for optimisation and provide a natural mechanistic procedure based on a neuro-symbolic multi-agent reinforcement learning approach, which leads to the emergence of numeral systems that are close to the Pareto frontier of efficiency between lexicon size and average morphosyntactic complexity.




\section{Meta-grammars for recursive number systems}
\label{sec:grammar}
The grammar introduced by \cite{Hurford-1975, Hurford1987LanguageAN} allows for the representation of numerals in natural languages. It relies on two sets: \emph{$D$} -- enumerating the set of lexicalised digits (e.g. one, two, three,...), and \emph{$M$} -- enumerating the set of multipliers (e.g. ten, hundred,...). Some examples are shown in Table \ref{tab:human_DsMs}.
We use a slightly modified version of this grammar, which is given in equation \eqref{eq:our-grammar}. Our modification removes the single $M$ from the construction of \textit{Phrase} leaving only $Num * M$ as an option. 
When optimising $D$ and $M$ pairs this small modification results in more natural, human-like systems as it captures that multipliers ought to have higher costs than digits.

\begin{equation}
\label{eq:our-grammar}
\begin{split}
&\mathbf{Num} = D\ | \ Phrase\ | \ Phrase + Num\ |\ Phrase - Num \\
&\mathbf{Phrase} = \ Num * M   
\end{split}
\end{equation}

This meta-grammar is a mutually recursive (\textit{Num} and \textit{Phrase} are defined in terms of each other) non-free datatype (not every number has a canonical representation). A specific (human) number system consists of a combination of sets $D$ and $M$, together with a commitment to a unique representation for each number where there are alternatives. That is, the same $D$ and $M$ pair can appear in several languages, differing in which representations are chosen for the numerals.

\begin{table}[htbp]
  \centering
  \caption{Some $D$ and $M$ pairs associated with different human numeral systems.}
  \label{tab:human_DsMs}
  \begin{tabular}{ccccccc}
    \toprule
    Language & $D$ & $M$ \\
    \midrule
    English & [1, 2, 3, 4, 5, 6, 7, 8, 9, 11] & [10] \\
    French & [1, 2, 3, 4, 5, 6, 7, 8, 9] & [10, 20] \\
    Kunama  & [1, 2, 3, 4] & [5, 10] \\
    \bottomrule
  \end{tabular}
\end{table}
 


\subsection{Complexity Metrics}
Optimization with the meta-grammar constitutes choosing appropriate sets $D$ and $M$, and then the composition of the remaining numerals using the constructions operators $(+, -, *)$. 

The \emph{morphosyntactic complexity} of a \textit{Num} is simply defined as the size, i.e. number of symbols, in the expression built using the meta-grammar. For example, the single digit ``2" has complexity 1 while ``2 * 10" has complexity 3 etc. This measure is referred to as $ms\_complexity$.

Recall that the meta-grammars allows non-canonical representation of numerals. For a given pair of $D$ and $M$, a set of concrete languages, $\mathcal{L} = \{L_{1}, \dots, L_{N}\}$, can be induced. For each concrete language, the average morphosyntatic complexity under the need distribution can be calculated according to Equation \ref{eq:avg_morpho_complexity}. The need distribution is parameterised as $P(n) \propto n^{-2}$ which captures that smaller numbers are typically used more frequently and thus better expressed with low complexity constructs \citep{dehaene1992cross, Xu2020}. The evaluation is limited to natural numbers from 1 to 99, as numbers beyond this range have a very low probability under the given distribution, making their contribution to the average morphosyntactic complexity negligible.

\begin{equation}
\label{eq:avg_morpho_complexity}
   avg\_ms\_complexity(L) = \sum_{n\in[1, \dots, 99]}  P(n) \cdot ms\_complexity(n, L) \\
\end{equation}

 We refer to the language which achieves the most compact representation 
 as $L_{min}$: 

\begin{equation}
    L_{min} = \arg\min_{L\in \mathcal{L}}(avg\_ms\_complexity(L))
\end{equation}

$L_{min}$ is the lower-bound on the average morphosyntactic complexity obtainable by a given $D$ and $M$ pair. We note that when optimising $D$ and $M$ pairs we are interested in minimising the measure shown in Equation \ref{eq:morpho_complex}.

\begin{equation}
\label{eq:morpho_complex}
    avg\_ms\_complexity(D, M) = avg\_ms\_complexity(L_{min})
\end{equation}

The minimum languages associated with the final population of $D$ and $M$ pairs will provide an approximation to the set of optimal languages that lie on the Pareto frontier. Languages in this set are optimal in the sense that a reduction in average morphosyntactic complexity cannot be obtained without an increase in lexicon size and vice versa. 


\section{Reinforcement Learning for Grammars}

\begin{figure}[t]
    \centering
    \includegraphics[width=0.45\textwidth]{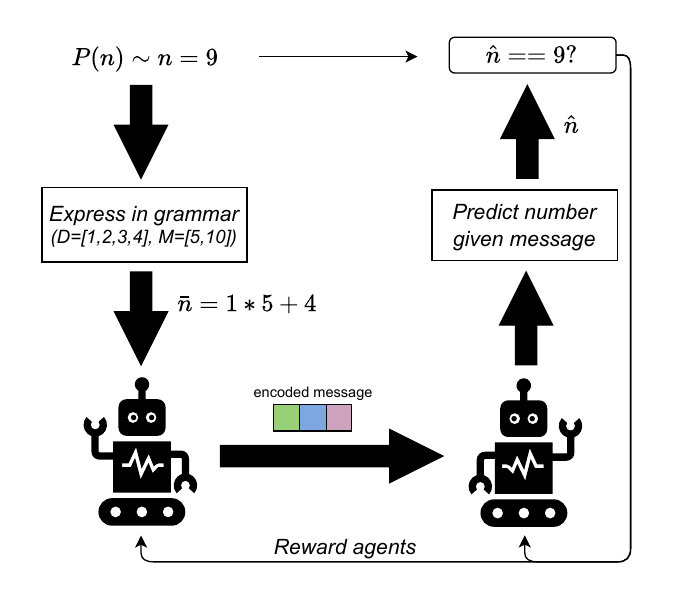}
    \caption{An example of a communication round: a number, e.g. $n=9$, is sampled from the need distribution. The corresponding representation in terms of the current $(D,M)$ is passed to the speaker, which encodes it neurally. The listener receives this message and outputs the number it thinks it refers to. Both agents are rewarded if the guess is correct.}
    \label{fig:rl_agents_example}
\end{figure}

Like in previous work by \citet{carlsson2021numeral} we study the emergence of number systems as a two-agent Lewis signaling game \citep{Lewis1969} using reinforcement learning. One agent, the speaker, has to convey a numerosity encoded as a neural message, and is rewarded if the listener can reconstruct the number. 
\citet{carlsson2021numeral} left the extension to recursive numerals systems open for future work.
Their agents were only capable of sending single messages, which is incompatible with the compositional nature of recursive numeral systems, and it was also unclear how to integrate a grammar structure like \citet{Hurford-1975}'s. 
To allow the agents to manipulate discrete grammars and send variable-length messages, we use Hurford's grammar (Equation \ref{eq:our-grammar}) to construct variable-length symbolic expressions that are used as input data for the speaker. These are then communicated neurally with variable-length messages, using an LSTM architecture \citep{hochreiter1997long} and reinforcement learning. The length of the speaker’s messages depends on the length of the expression of the numeral being communicated, given the current contents of D and M. The listener receives this message and outputs the number it believes it refers to. Figure \ref{fig:rl_agents_example} shows an example round of our framework. 

\begin{algorithm}[hbt!]
\caption{Learning Efficient Recursive Numeral Systems}\label{alg:rec}
\KwResult{Final $(D,M)$ pair}

\textbf{Requires}: initial $(D_{curr}, M_{curr})$ pair, randomly initialized speaker $\mathcal{S}$ and listener $\mathcal{L}$;

Compute current dataset $L_{min, curr}$\;

Pre-train $\mathcal{S}$ and $\mathcal{L}$ on $L_{min, curr}$ via communication\;

\For{$i \in [1, \dots, \text{max steps}]$}
{  
    \textit{\textbf{Grammar Modification Phase}}
    
    Perform mutations to $(D_{curr}, M_{curr})$ from Table \ref{tab:actions}\;

    Randomly sample $(D_{alt}, M_{alt})$\;

    Initialize $\text{Q-values}=[0.0, 0.0]$\;

    \textit{\textbf{Communication Phase}}
    
    \For{$j \in [1, \dots, \text{max iterations}]$}
    {

    Sample a value $n$ between $0.0$ and $1.0$\;
    \eIf{$1 - \epsilon <= n$}
    {
        Select highest \text{Q-value} grammar\;
    }
    {Select the least explored grammar\;}

        Sample batch from picked grammar $G_k$=($D_k$,$M_k$) where $k \in \{curr, alt\}$;

        Train $\mathcal{S}$ and $\mathcal{L}$ via communication on batch;

        $Q(G_k) \leftarrow Q(G_k) + \alpha ( r - Q(G_k))$;
 
    } 

    Update $(D_{curr}, M_{curr})$ based on highest Q-value;
    
}
\end{algorithm}

Algorithm \ref{alg:rec} summarises our method for learning recursive number systems via RL communication. It takes a starting grammar ($D_0$, $M_0$) as input. From this, we compute $L_{min,0}$, i.e. the set of minimal representations for each numeral in the range $1-50$, increasing it from \citet{carlsson2021numeral} who had a range of $1$ to $20$. The agents then use these variable-length expressions to play a signaling game: the speaker, given a numerical quantity sampled from the need distribution,  retrieves the corresponding expression in $L_{min,0}$. The speaker then neurally encodes a message given this expression. The listener receives this message and then outputs the number they think the speaker is trying to represent. The agents start \textit{tabula rasa}, i.e. are randomly initialized, and are jointly trained together via reinforcement learning: they are rewarded positively only if the listener predicts the right numerical quantity from the message communicated by the speaker. Note that in our setup only the speaker has access to the grammar, while the listener has it indirectly through the messages produced by the speaker, which are random at the start but will obtain structure through successful communication.

After some rounds the agents stop communicating, allowing the speaker to consider modifying the current grammar ($D_{curr}$, $M_{curr}$), with the aim of exploring possibly more efficient versions. 
We propose a small set of six conservative modifications that can be applied to ($D_{curr}$, $M_{curr}$), given in Table \ref{tab:actions}. 
Through only applying small changes, the language shifts gradually allowing the listener to leverage some of what it has learned from previous interactions rather than starting from scratch at each iteration. 

\begin{table}[h]
    \centering
    \caption{Our set of modifications and their descriptions.}
    \begin{tabular}{c p{6cm}}
        \hline
        \textbf{Action} & \textbf{Description} \\
        \hline
        $m_0$ & Add highest numeral not in $D$ or $M$ to $D$ \\
        $m_1$ & Add highest numeral not in $D$ or $M$ to $M$ \\
        $m_2$ & Move lowest numeral in $M$ to $D$ \\
        $m_3$ & Move highest numeral in $D$ to $M$ \\
        $m_4$ & Remove highest numeral in $D$ \\
        $m_5$ & Remove highest numeral in $M$ \\
        \hline
    \end{tabular}
    \label{tab:actions}
\end{table}

We randomly sample one of these alternatives ($D_{alt}$, $M_{alt}$), deriving a new language and the resulting alternative minimal expressions $L_{min, alt}$. This new alternative language allows us to set up recursive numeral systems-learning as a competitive game between the two grammars: at each step of communication, the speaker will sample expressions from one of the two grammars, and will then collect feedback on how successful they are at communicating numerical quantities with each of them. 
To choose betwen the two grammars, we employ a \emph{bandit} algorithm \citep{lattimore2020bandit} (which are a class of algorithms for sequential decision making modelling the choice between arms of a slot machine as an analogy). The two number systems correspond to the bandit's arms, which the speaker interacts with to learn which one generates the expressions that give the best reward in terms of communicative success. 
More specifically, the speaker stores and updates two Q-values that estimate the ``goodness'' of both grammars. The Q-value is updated at each communication step with the rewards collected when using the corresponding language, as $Q(G_i) \leftarrow Q(G_i) + \alpha ( r - Q(G_i))$, where $\alpha$ is the learning rate. The Q-values are used to pick what grammar to choose: with probability $\epsilon$ the highest valued grammar is picked, otherwise the least used one is. This is to balance the \textit{exploration-exploitation} problem, which means exploring new under-used actions while also exploiting the knowledge accumulated via the previous trials. At the end of communication, we set the new current grammar ($D_{curr}$, $M_{curr}$) to the grammar with the highest Q-value. The process then recurse until the maximum number of iterations is reached. 

\paragraph{Implementation Details}
We parametrise the two agents policies as LSTMs \citep{hochreiter1997long} with one embedding layer of size $5$ and a single LSTM layer of hidden size $100$. The speaker policy maps from expressions of variable length to messages of the same length. The listener policy maps from variable length messages to the space of numerals considered, predicting directly the numerical quantity. We train both policies jointly using the REINFORCE loss \citep{williams1992simple} via two ADAM optimizers \citep{kingma2015adam} with learning rates of $0.002$ using a batch size of $32$. We additionally add an entropy term to the listener loss to encourage exploration.
The default value of the bandit's learning rate is $\alpha=0.1$ and $\epsilon=0.2$, while the Q-values are initialized everytime a new round of communication starts to $0.0$. Each communication round lasts for $40000$ epochs. For each modification to the grammar, we first check that the resulting new grammar can cover all the numbers in the range we are interested in, otherwise we discard it and thus it cannot be sampled as the alternative grammar.

\section{Results and Discussion}

\subsection{Communication Leads to More Efficient Languages}

\begin{figure}[t]
    \centering
    \includegraphics[width=0.48\textwidth]{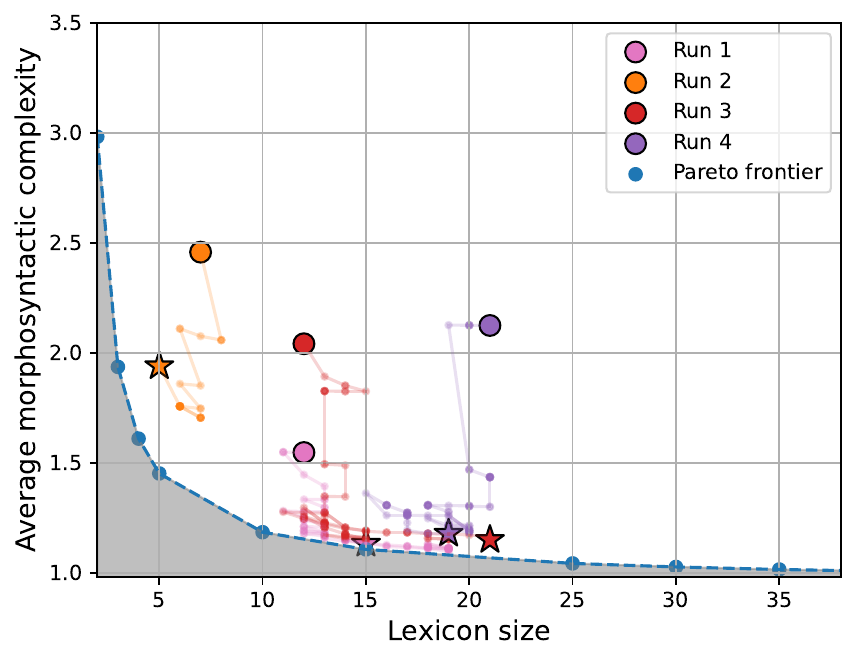}
    \caption{Trajectories showing the evolution of the languages during communication between our agents,  starting from points 1-5 in Table \ref{tab:starting_points} (for visual clarity not all trajectories are included). The agents tend towards languages that are closer to the Pareto frontier. The grey area represent unobtainable configurations.}
    \label{fig:rl_optimization}
\end{figure}

\begin{figure}[t]
    \centering
    \includegraphics[width=0.48\textwidth]{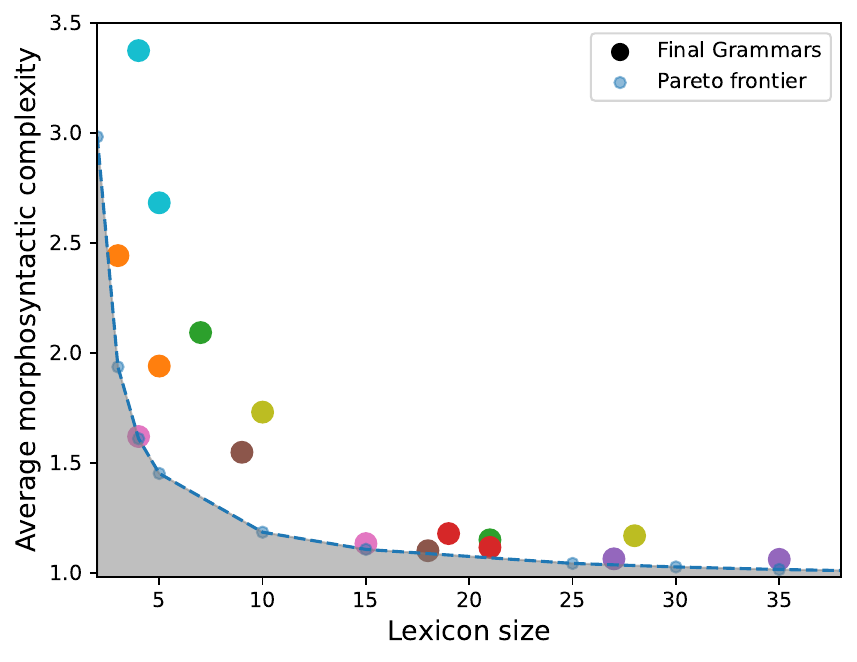}
    \caption{Final languages of our agents, plotted in terms of lexicon size and average morphosyntactic complexity. The final grammars tend to lie close to the Pareto frontier. Points are colour coded based on their starting $(D,M)$ pair.}
    \label{fig:rl_final_languages}
\end{figure}

In Figures \ref{fig:rl_optimization}-\ref{fig:rl_final_languages}, we show that our neuro-symbolic method results in ($D$, $M$) pairs that lie very close to the Pareto frontier, just like human numeral systems. 
The results are obtained by running Algorithm \ref{alg:rec} for $100$ iteration, starting from the ($D$, $M$)-configurations in Table \ref{tab:starting_points}. These are randomly selected among non-Pareto optimal ($D$, $M$)-combinations representing languages covering the entire numeral range between $1$ and $50$. 
We perform $2$ runs for each such starting point (Figure \ref{fig:rl_final_languages}), showing that we can obtain different end languages from the same starting point, but always tend towards the Pareto frontier.  

\begin{table}[htbp]
  \centering
  \scriptsize 
  \caption{Starting $D$ and $M$ pairs used in RL experiments.}
  \label{tab:starting_points}
  \begin{tabularx}{\linewidth}{l|ll}
    \toprule
    \textbf{\#} & \textbf{$D$} & \textbf{$M$} \\
    \midrule 1 &
    {[1, 2, 3, 20, 35, 37, 40, 47, 49]} & {[4, 25, 45]} \\
    2 & {[1, 4, 7, 8, 15]} & {[10, 33]} \\
    
    3 & {[1, 20, 25, 28, 31, 39, 41, 45]} & {[2,3,4,15]} \\
    4 & {[1, 4, 19, 21, 39, 40, 45, 47, 49]} & {[3,5,8,10,18,23,28,30,37,42,43,48]} \\
    5 & {[1, 2, 3, 4, 5, 10, 20, 30, 35, 40, 47, 49]} & {[6, 11, 13, 15, 45, 50]} \\
    6 & {[1, 2, 35, 37, 40, 47, 49]} & {[3, 5, 10, 20, 30, 40]} \\
    7 & {[1, 4, 12]} & {[9, 25]} \\
    8 & {[1, 4, 17, 22, 49]} & {[9, 10, 25, 28, 31, 41, 45]} \\
    \bottomrule
  \end{tabularx}
\end{table}

The trajectories in \ref{fig:rl_optimization} show that our agents tend to consistently prefer more efficient languages, which over time move closer and closer to the Pareto frontier. This demonstrates that the pressures for efficient communication that come into play in our two-agents game produce languages that are more efficient in terms of tradeoff between average morphosyntactic complexity and lexicon size. These more efficient languages are easier to learn for our agents, connecting our results to works on the link between simplicity and ease of learning \citep{steinert2020ease}.

Our final languages in Figure \ref{fig:rl_final_languages} cover a wider span along the Pareto frontier than human languages, which tend to have a lexicon size of around 10-15. This might depend on our random starting points, or not-yet captured human constraints leading to a preference for smaller lexicon sizes.  
The need distribution might also influence our results: adopting the exponential need distribution used in \citet{denicRecursiveNumerals} for the calculation of the average morphosyntactic complexity, we note that larger terms become very rare when it is used to sample numerical quantities for our agents to communicate. Further work with language evolution paradigms such as iterated learning \citep{smith2003iterated} might solve this issue. 
We also note that some starting configurations that are further away from the Pareto frontier may require more iterations to fully reach the frontier. We kept the number of steps to $100$ for consistency sake. Still, our experiments show a consistent trend towards the Pareto frontier.

To summarise, our approach shows that more efficient languages are preferred, which provides a simple mechanistic explanation of why human recursive numeral systems developed to be efficient in the tradeoff between average morphosyntactic complexity and lexicon size.

\subsection{Reward Structure for Numeral Systems}

Compared to approximate and exact restricted numeral systems, where a word may refer to a range of numerosities, recursive numeral systems refer to exact quantities. Therefore,
the proximity-based reward function used in \citet{carlsson2021numeral}, which promotes guesses that are closer to the target value, is not suitable here. We instead reward our agents via an exact reward, i.e. only if the listener predicts the correct numerosity. This in turn makes the multi-agent reinforcement learning problem much harder, as reward is \textit{sparser}. Adapting larger LSTMs and using a larger batch size handled this.

\subsection{Characteristics of Pareto Optimal Languages}

\begin{figure}[t]
    \centering
    \includegraphics[width=0.45\textwidth]{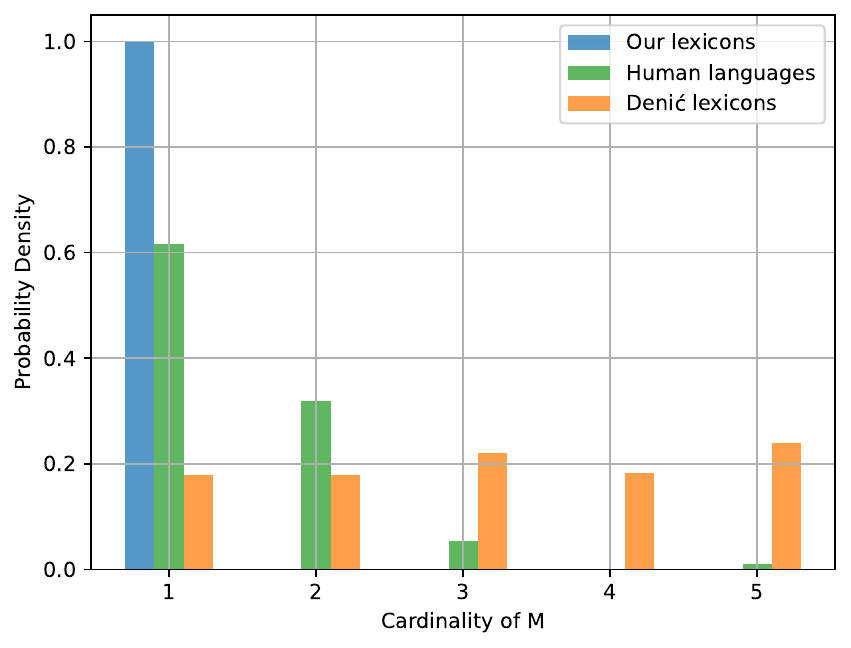}
    \caption{Comparison of lexicons induced by our meta-grammar and Hurford's in terms of the cardinality of $M$.}
    \label{fig:compare_m}
\end{figure}

We motivate our change of meta-grammar by comparing the cardinality of $M$ for languages on the Pareto frontier (Figure \ref{fig:compare_m}). This shows that our meta-grammar results in optimal languages which contain a single multiplier in $M$. Instead, with Hurford's original meta-grammar, the optimal languages show an approximately uniformly distribution across cardinalities of $M$ within the permitted range. Human languages tend to have grammars with just a few multipliers in M: numeral systems with one or two multipliers are mostly common. Although our resulting lexicons are not exactly matching this, they have a consistent bias towards fewer multipliers, which was not the case if we used Hurford's meta-grammar. Furthermore, it is notable that the genetic algorithm utilised by \citet{denicRecursiveNumerals} imposes an artificial constraint upon the meta-grammar which restricts it to systems with no more than five elements in $M$ with the intention of better representing human systems. Without the constraint, optimal lexicons would mostly have an even higher cardinality of $M$. Instead, our modification makes it so this hard constraint is not needed when finding the Pareto frontier.


Table \ref{tab:compare_language}, shows examples of Pareto-optimal languages and how they differ in their composition of $D$s and $M$s. The example deduced from Hurford's original grammar results in an $L_{min}$ which does not have a predictable recursive structure, as the expressions for close numbers vary a lot in terms of form and Ms used as basis. This is in contrast to ours and a human language from the Mixtec group (see  \citet{denicRecursiveNumerals}, appendix) referred to as \textit{Type 4-MixtecA}. We suggest that our meta-grammar, when used for optimization, results in lexicons that do not overrely on Ms, like with natural human languages. A downside of our approach is that when representing natural numeral systems with our modified meta-grammar some nuances get lost, e.g. the numeral 10 in English would have to be encoded as ``1*10" with our modified meta-grammar. While this is satisfactory for the optimization purposes of our current work aims, future work should try to find a grammar variant that is both closer to natural languages and work for optimization. 

\begin{table}[htbp]
  \centering
  \caption{A comparison of different $D$ and $M$ pairs associated with Pareto-optimal languages with a lexicon size of $12$.}
  \label{tab:compare_language}
  \begin{tabular}{ccccccc}
    \toprule
    Grammar & $D$ & $M$ \\
    \midrule
    Hurford & [1, 2, 3, 5, 6, 9, 10, 11, 14] & [4, 7, 25] \\
    \eqref{eq:our-grammar} (ours) & [1, 2, 3, 4, 5, 6, 7, 8, 9, 10, 11] & [12] \\
    4-MixtecA  & [1, 2, 3, 4, 5, 6, 7, 8, 9] & [10, 15, 20] \\
    \bottomrule
  \end{tabular}
\end{table}


\section{Conclusions and Future Work}


Our neuro-symbolic method shows how efficient recursive numeral systems develop as a trade-off between morphosyntactic complexity and lexicon size, 
through a combination of two-agent signaling games, reinforcement learning and symbolic language modifications.
A minor modification to Hurford's well-established meta-grammar for expressing numeral systems made it more suitable for optimisation via RL. 
Our contributions provide a mechanistic explanations for the emergence of efficient recursive numeral systems, supporting the work of \cite{denicRecursiveNumerals}.
In future work, we intend to explore how cultural transmission paradigms like iterated learning might influence the languages our agents develop, possibly leading to languages even closer to human ones in terms of average morphosyntactic complexity and lexicon size, as \citet{carlsson2024cultural} demonstrates for the colours domain and \citet{guo2019emergence} for numerical quantities. We also want to explore how different need distributions affect the efficiency of our agents' languages. Finally, we are interested in creating a unifying framework that can demonstrate how approximate, exact restricted and recursive numeral systems might emerge from the same system and possibly transition from one to the other through communication and inter-generational transmission.

\section{Acknowledgments}

This work was supported by funding from the Swedish Research Council and the Wallenberg AI, Autonomous Systems and Software Program
(WASP) funded by the Knut and Alice Wallenberg Foundation. The computations in this work were enabled by resources provided by
the Swedish National Infrastructure for Computing (SNIC).

\setlength{\bibleftmargin}{.125in}
\setlength{\bibindent}{-\bibleftmargin}

\bibliography{CogSci_Template}

\end{document}